# A List of Household Objects for Robotic Retrieval Prioritized by People with ALS (Version 092008)

Young Sang Choi[1], Travis Deyle[1], and Charles C. Kemp[1]
[1]Healthcare Robotics Lab, Georgia Institute of Technology, USA

*Abstract*—This technical report is designed to serve as a citable reference for the original prioritized object list that the Healthcare Robotics Lab at Georgia Tech released on its website in September of 2008. It is also expected to serve as the primary citable reference for the research associated with this list until the publication of a detailed, peer-reviewed paper.

The original prioritized list of object classes resulted from a needs assessment involving 8 motor-impaired patients with amyotrophic lateral sclerosis (ALS) and targeted, in-person interviews of 15 motor-impaired ALS patients. All of these participants were drawn from the Emory ALS Center.

The prioritized object list consists of 43 object classes ranked by how important the participants considered each class to be for retrieval by an assistive robot. We intend for this list to be used by researchers to inform the design and benchmarking of robotic systems, especially research related to autonomous mobile manipulation.

## I. Motivation & Introduction

Studies have consistently shown that object retrieval would be a valuable task for assistive robots to perform, yet detailed information about the needs of patients with respect to this valuable task has been lacking [1]. Physically impaired people often have difficulty retrieving objects from the floor and shelves, and these problems can be exacerbated by reduced strength and dexterity in the upper extremities, which can increase the chances of dropping an object.

Within this technical report, we present one of the results of our efforts to better understand the needs of physically impaired patients with ALS. Specifically, we present a prioritized list of object classes for retrieval by assistive robots. In addition to this written report, we have created a website with a matched list of objects that can be purchased through online vendors, so that researchers have the opportunity to more easily work with the same set of objects and compare results: http://www.hsi.gatech.edu/hrl/object_list_v092008.shtml.

We intend for this list to be used by robotics researchers to better inform their designs and methods of evaluation. For example, this list provides a guide to robot design since the set of objects has implications for specifications such as payload capacity. Additionally, the list suggests methods for benchmarking the object fetching capabilities of a robot, since a robot could be empirically evaluated using representative objects from the top N object categories from the prioritized list. For instance, there are 25 objects with an average ranking of "Important" to "Very Important" (i.e., in the range [6,7]) for assistive robotic retrieval.

This list also speaks to larger issues within the robotics research community. For example, unlike the speech and vision communities [2], [3], robotics researchers working on robot manipulation have yet to define common benchmarks by which they can evaluate performance in comparable ways.

## II. A Prioritized List of Household Objects

We initially performed a needs assessment with 8 ALS patients for which we provided patients and caregivers with cameras and notepads to document for one week when objects were dropped or were otherwise unreachable in daily life. After this time, we interviewed each patient in his or her home. Based on this initial study, we created an interview questionnaire to assess the importance of various objects for robotic retrieval using the Likert scale [4]. We administered this survey to 15 patients through in-person, interviews that were each approximately 30 minutes long.

By directly asking patients about their object retrieval priorities, we were able to implicitly address a host of important factors including frequency of use, difficulty of retrieval, object importance, and personal preference. We conducted in-person interviews based on the recommendations of experts at the Emory ALS Clinic, who believed that this would be the only way to ensure high-quality results.

It is worth emphasizing that even though the questionnaire used images of specific objects, the list represents *classes* of objects rather than specific instances (i.e. the object class of all shoes, rather than a particular brand, style, or color of shoe).

### A. Interview Methods

In total, 15 ALS patients (demographics in Table I) participated in the interviews. We recruited patients during the course of regular medical examinations and consultations at the Emory ALS Clinic. Patients were pre-screened by nursing professionals, and depending on their interest, were introduced to the researchers. Interviews began by briefly introducing the goals and existing functionality of our lab's mobile manipulator, EL-E, before proceeding to the interview proper. This provided patients with a sense of how an assistive object-fetching robot might look and function.

The interview began with the interviewer reading consent forms to the participants and receiving their signatures. When a participant had difficulty in writing, a caregiver signed on behalf of the participant and a demographic survey followed.

In accordance with the IRB approved protocol, the interviewer asked participants to indicate the relative importance of each object in a list (with the representative pictures in

TABLE I
DEMOGRAPHIC INFORMATION FOR INTERVIEW PARTICIPANTS

| | |
|---|---|
| Gender | Male (10), Female (5) |
| Ethnicity | White (7), African American (8) |
| Age | 42 - 81 (avg 58.1) years |
| Diagnosis Duration | 3 - 87 (avg 30.4) months |

Table II) based on a 7-point Likert scale. The text from the questionnaire and the details of the protocol follow.

> For a research project at Georgia Tech and Emory, we are developing a robot to help people to manipulate everyday objects. We are trying to find a list of common objects to be useful in robot manipulation research for us and other robot researchers. Your help from experience would be essential in creating a validated list.
> Following is a list of objects with pictures which might be important to be retrieved by a robot if they are dropped or unreachable in your daily lives. For each object in the list, please give a number from 1 to 7 by following criteria for the importance of retrieval based on your experiences.
>
> | | |
> |---|---|
> | 7 | Very Important |
> | 6 | Important |
> | 5 | Slightly Important |
> | 4 | Neutral |
> | 3 | Slightly Unimportant |
> | 2 | Unimportant |
> | 1 | Very Unimportant |

The interviewer read these instructions to the participant and explained how to rate the relative importance of object retrieval. Then the interviewer showed the images of the objects printed on the questionnaire and requested the participant's rating on the 7 point Likert scale shown above. The participant would either verbally indicate the rating, or point to it if speech was too difficult.

An open-ended followup question concluded the interview.

> In your experience, if you have objects which were not included in the above list but you think are necessary to be retrieved by a robot, please list them.

After the first 10 interviews we performed a preliminary analysis of the results, and discovered that half of the participants inquired about articles of clothing such as socks, pants, and shirts. Based on the needs assessment, we had omitted clothing from the list because we expected ALS patients to have great difficulty dressing, even if clothes were retrieved by a caregiver (or robot). At the behest of participants, however, we added clothing items to the list for the final 5 interviews.

### B. Results: The Prioritized List of Objects

By averaging the Likert-scale rating of each object across all participants we computed a numerical ranking of the objects. Based on the responses of ALS patients, we can consider highly ranked objects to be more relevant (and broadly applicable) for robotically assisted object retrieval compared to lower ranked objects. The results of this ranking (with the averaged Likert score) is shown in Table II. We also selected a typical example from each object class in order to assign an approximate mass and longest length to the class for analysis. These object properties are recorded in the table as "weight" and "max size".

Among the 43 total objects, 40 objects were rated by all 15 participants. Three additional objects (socks, shirt, and pants) were added based on consensus about "additional objects" from the first 10 participants. While some participants thought the list was comprehensive, others suggested additional objects through the open-ended follow-up question, though there was no consensus on further omissions. Examples of the other objects mentioned were glass cups, milk jugs, coffee pots, tissues, and bath towels. Two patients also mentioned "*myself*" as an additional object, which represents the desire of some patients to have a robot capable of repositioning their bodies. Robots that can grasp and change the position of body parts (legs and arms) or the whole body would be extremely useful for this patient group. Although important, we have treated this as a separate category of manipulation that is distinct from object retrieval.

### III. CONCLUSION

We have proposed a ranked list of 43 everyday objects for the evaluation of assistive manipulation systems operating in domestic settings. By developing benchmarks tailored to specific application domains and user populations, robotics researchers have the opportunity to ground their research and answer the otherwise subjective question: *What is functionality is important?*. We also believe benchmarks of this nature can enable researchers without direct access to user populations to contribute to progress in a validated way.

### IV. ACKNOWLEDGMENTS

We thank the participants and caregivers for their time and insights. Recruitment of these participants was only possible due to help from Meraida Polak and Crystal Richards of the Emory ALS Center. We also thank Cressel Anderson, Hai Nguyen, Zhe Xu, and Zach Hughes for helping with this study. This research was supported in part by NSF grant IIS-0705130.

TABLE II
PRIORITIZED LIST OF OBJECTS

| Rank | Object Class | Image | Rating Mean | Rating Stdev. | Weight (grams) | Max size (cm) | Rank | Object Class | Image | Rating Mean | Rating Stdev. | Weight (grams) | Max size (cm) |
|---|---|---|---|---|---|---|---|---|---|---|---|---|---|
| 1 | TV Remote | 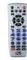 | 6.73 | 0.70 | 90 | 18 | 22 | Shoe | 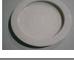 | 5.13 | 2.13 | 372 | 30 |
| 2 | Medicine Pill | 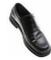 | 6.47 | 1.55 | 1 | 2.2 | 22 | Pen / Pencil | 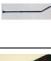 | 5.13 | 2.20 | 3 | 14 |
| 3 | Prescription Bottle | 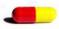 | 6.27 | 1.58 | 25 | 7 | 25 | Medicine Box | 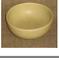 | 5.00 | 1.85 | 25 | 10 |
| 4 | Glasses | 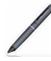 | 6.13 | 1.19 | 23 | 14 | 26 | Plastic container | 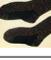 | 4.93 | 1.94 | 49 | 13 |
| 5 | Cordless Phone | 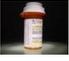 | 6.07 | 1.62 | 117 | 15 | 26 | Credit Card | 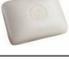 | 4.93 | 2.28 | 5 | 8.5 |
| 6 | Toothbrush | 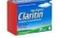 | 5.93 | 2.09 | 15 | 19 | 28 | Coin | 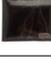 | 4.73 | 2.43 | 6 | 2.5 |
| 7 | Fork | 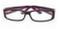 | 5.87 | 1.19 | 39 | 18 | 29 | Small Pillow | 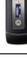 | 4.67 | 2.02 | 240 | 38 |
| 8 | Spoon | 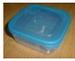 | 5.73 | 1.16 | 38 | 17 | 30 | Pants | 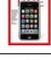 | 4.60 | 2.51 | 539 | 100 |
| 8 | Disposable bottle | 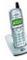 | 5.73 | 1.44 | 500 | 13 | 30 | Shirt | 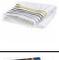 | 4.60 | 2.51 | 229 | 66 |
| 8 | Toothpaste | 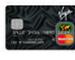 | 5.73 | 2.12 | 160 | 20 | 32 | Hairbrush | 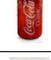 | 4.53 | 2.59 | 100 | 24 |
| 11 | Cup / Mug | 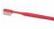 | 5.67 | 1.76 | 267 | 12 | 33 | Non-disposable bottle | 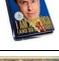 | 4.40 | 1.96 | 709 | 20 |
| 11 | Dish Plate | 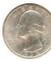 | 5.67 | 1.54 | 182 | 18 | 33 | Walking Cane | 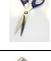 | 4.40 | 2.16 | 1140 | 94 |
| 11 | Dish Bowl | 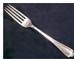 | 5.67 | 1.54 | 154 | 13 | 33 | Socks | 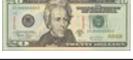 | 4.40 | 2.41 | 41 | 23 |
| 14 | Soap | 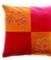 | 5.53 | 2.42 | 116 | 9.5 | 36 | Wallet | 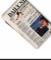 | 4.33 | 2.35 | 116 | 100 |
| 14 | Cell Phone | 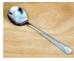 | 5.53 | 2.03 | 76 | 9 | 37 | Magazine | 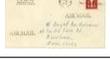 | 4.27 | 2.05 | 206 | 27.5 |
| 14 | Hand Towel | 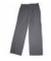 | 5.53 | 1.73 | 65 | 58 | 38 | Can | 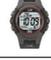 | 3.93 | 2.02 | 350 | 6.4 |
| 14 | Book | 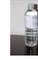 | 5.53 | 1.41 | 532 | 24 | 39 | Scissors | 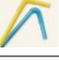 | 3.87 | 2.45 | 25 | 14 |
| 18 | Bill | 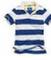 | 5.47 | 2.07 | 1 | 13.5 | 39 | Newspaper | 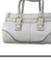 | 3.87 | 2.10 | 247 | 31 |
| 18 | Mail | 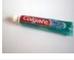 | 5.47 | 2.23 | 22 | 24 | 41 | Wrist Watch | 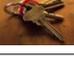 | 3.53 | 2.42 | 86 | 10 |
| 20 | Straw | 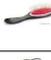 | 5.40 | 1.64 | 1 | 20 | 42 | Purse / Handbag | 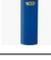 | 2.93 | 2.31 | 380 | 24 |
| 21 | Keys | 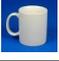 | 5.27 | 2.40 | 24 | 8.5 | 43 | Lighter | 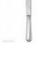 | 2.73 | 2.34 | 91 | 6 |
| 22 | Table Knife | 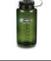 | 5.13 | 1.68 | 76 | 24 | | | | | | | |